\documentclass[12pt]{article}
\usepackage{geometry}
\usepackage{xcolor}
\usepackage{fancyhdr}
\usepackage[labelfont=bf]{caption}
\usepackage{graphicx}
\usepackage{float}
\usepackage{booktabs}
\usepackage{tabularx}
\usepackage{titling}
\usepackage{titlesec}
\usepackage{subfig}
\titleformat{\section}{\normalfont\bfseries\filcenter}{}{0pt}{}
\titleformat{\subsection}{\normalfont\bfseries\filcenter}{}{0pt}{\itshape}
\titleformat{\subsubsection}{\normalfont\bfseries\filcenter}{}{0pt}{\itshape}
\titlespacing*{\section}
  {0pt}{-.1\baselineskip}{-.1\baselineskip}  
\titlespacing*{\subsection}
   {0pt}{-.1\baselineskip}{-.1\baselineskip}  

\usepackage[]{cite}
\usepackage[hidelinks]{hyperref} 
\usepackage{authblk} 

\usepackage{lipsum}
\usepackage[normalem]{ulem}

\usepackage[T1]{fontenc}
\date{}
\providecommand{\keywords}[1]
{
   \small	
  \textit{\hspace{-1em} Keywords: } #1
}

\title{High Throughput Soybean Pod-Counting with In-Field Robotic Data Collection and Machine-Vision Based Data Analysis}

\author[1]{Michael McGuire}
\author[1]{Chinmay Soman}
\author[2]{Brian Diers}
\author[3,1]{Girish Chowdhary}
\affil[1]{EarthSense Inc., USA}
\affil[2]{University of Illinois at Urbana Champaign USA, Crop Sciences}
\affil[3]{University of Illinois at Urbana Champaign USA, Agricultural and Biological Engineering and Computer Science}

\begin{document}
\maketitle
\thispagestyle{fancy}
\begin{abstract}
  \noindent  We report promising results for high-throughput on-field soybean pod count with small mobile robots and machine-vision algorithms. Our results show that the machine-vision based soybean pod counts are strongly correlated with soybean yield. While pod counts has a strong correlation with soybean yield, pod counting is extremely labor intensive, and has been difficult to automate. Our results establish that an autonomous robot equipped with vision sensors can autonomously collect soybean data at maturity. Machine-vision algorithms can be used to estimate pod-counts across a large diversity panel planted across experimental units (EUs, or plots) in a high-throughput, automated manner.  We report a correlation of 0.67 between our automated pod counts and soybean yield. The data was collected in an experiment consisting of 1463 single-row plots maintained by the University of Illinois soybean breeding program during the 2020 growing season. We also report a correlation of 0.88 between automated pod counts and manual pod counts over a smaller data set of 16 plots.
  
\end{abstract}

\keywords{\textit{high throughput phenotyping, robotics, soybean} \vspace{8ex}}

\section{Introduction}

Increasing yield is the most important breeding objective of most soybean breeders. In research plots, soybean yields are quantified  by harvesting each plot individually, a labor intensive process. Breeders also estimate yield visually, but it is difficult to accurately identify the high-yielding lines this way. Soybean yield is highly correlated to the number of pods on plants but manually counting pods is also very labor intensive\cite{riera2020deep,agriculture10080348,li2019soybean,yu2016development}. Accurately and efficiently estimating the yield of experimental lines is thus a key bottleneck in soybean breeding pipelines, especially at the early evaluation stage where a large number of lines need to be evaluated. Here we report the results for pod counting from in-field imagery using "TerraSentia"–a robotic, high-throughput field phenotyping system that does not utilize any destructive sampling. Efficiently identifying high performing varieties early in the breeding pipeline will reduce costs and increase genetic gains for new lines. In the near future, the results presented here indicate the potential for direct yield estimates in breeding as well as production fields. Our results demonstrate a clear potential for low-cost high-throughput phenotyping with robots as the solution the phenotyping bottleneck\cite{furbank2011phenomics}.


Robotic and machine-learning based in-field pod counting has been a challenging problem.  First, the visual data collected by the robot must be thorough and reliable, and the data processing pipeline must associate the data to specific experimental units (plots) without any errors. Second, pods must be detected within this data with minimal false negatives and false positives. Third, care must be taken to  model the soybean plants and to distinguish unique pods from each other as much as possible, ensuring that the algorithms deliver a sampling of in-field data without  double counting. Any problem in the data collection scheme will delay development of the detection and spatial reasoning of the soybean plant and pods. Likewise, discovering limitations in detection or geometric modeling of the plot may require adding or improving sensors on the recording device, limiting the usefulness of data collected in past years. The limitation of data collection and experimentation to a brief window once each year is an additional complicating factor.

In this report we describe the results from a field-phenotyping system specifically designed to address these challenges. While there has been some interest in automated counting of soybean pods \cite{riera2020deep,agriculture10080348,li2019soybean,yu2016development}, an end-to-end system capable of addressing the challenges described above and delivering high-throughput pheonotyping information in an efficient manner has not yet been reported.


\section{Related Work}

Prior approaches to estimating soybean pod count or yields in an automated manner have made use of special vehicles for gathering data as well as computer vision techniques to detect soybean pods in images. However, a complete high-throughput pipeline for predicting yield from a large number of plots has not yet been reported. Li et al \cite{li2019soybean} made a public dataset of soybean images by marking the center of each seed in a soybean pod and trained a detector on this data for soybean seed counting. However, their dataset contained soybean pods scattered on a black background in a lab setting, rather than in a real-world field setting. Wei et al \cite{agriculture10080348} showed that there is a strong correlation between soybean yield and soybean pod count in public datasets, indicating that a soybean pod count algorithm can be a valid proxy for estimating yield, and they experimented with image processing and computer vision for detecting soybean pods. Yu et al \cite{yu2016development} used an Unmanned Aerial Vehicle (UAV) to measure crop geometric features and predict final yield with strong correlation. Finally, Riera et al \cite{riera2020deep} used a tracked robot with a mounted Logitech C920 camera to collect data on a limited number of plants, and a RetinaNet model for soybean pod detection.

\section{Materials and Methods}

\subsection{TerraSentia Field Phenotyping System}

\begin{figure}
    \centering
    \includegraphics[width=.9\textwidth]{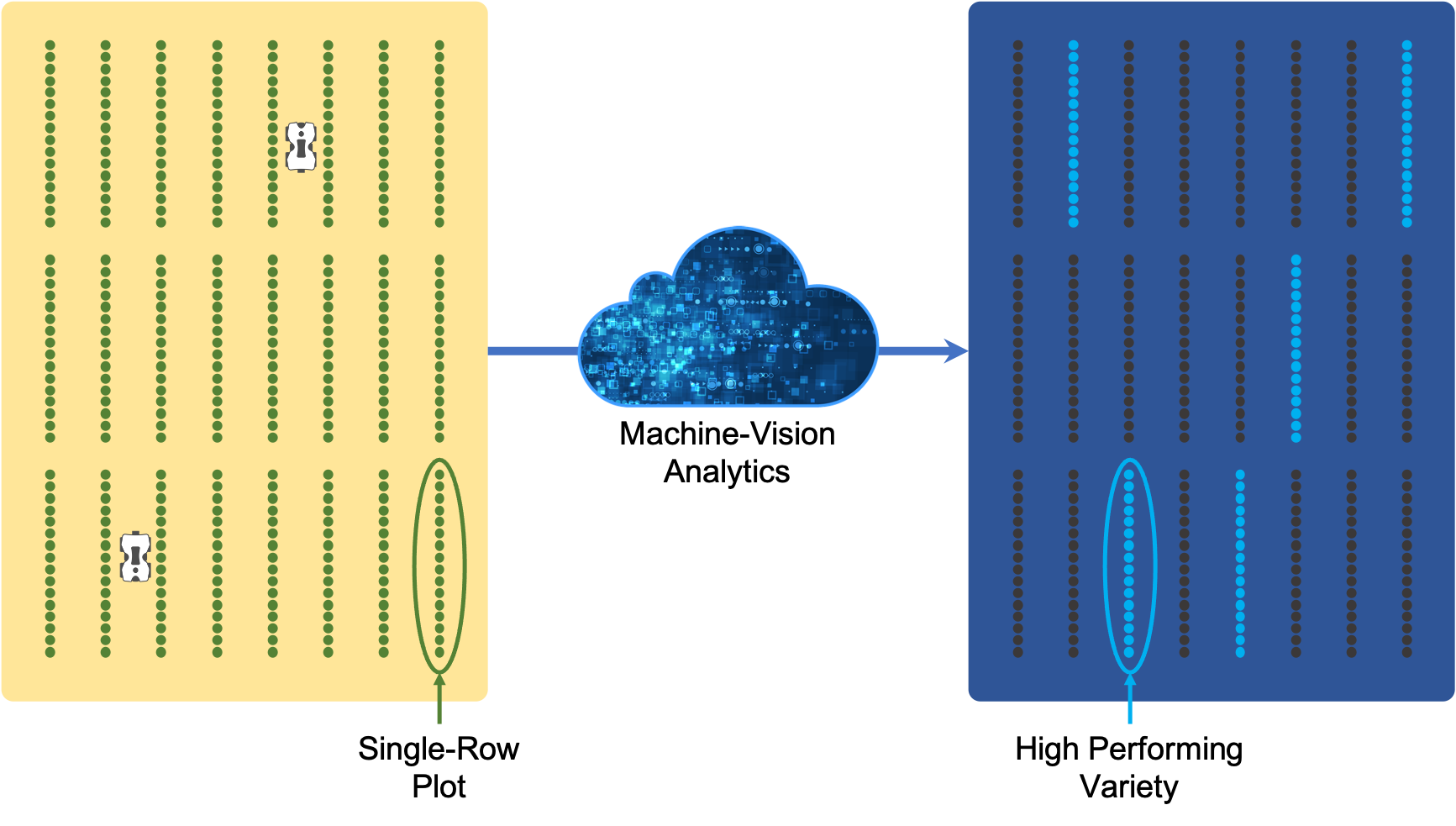}
    \caption{A visualization of the high-throughput phenotyping pipeline, shown here with single-row plots as utilized in this report. Robots collect data in an autonomous manner through breeding diversity panels, split into plots or experimental units. The data is transmitted to the cloud (or a remote computer at the Edge) where it is processed, yield estimates are computed, and reported back to the breeder with a user-friendly interface.}
    \label{fig:ES_pheno}
\end{figure}

The TerraSentia high-throughput phenotyping system consists of the TerraSentia robots that collect field data with multiple sensors, a user-interface for programming the robots, and sensor-fusion and analytics algorithms that extract quantitative trait information from the raw data. The robot can traverse autonomously using a combination of GPS, LiDAR, or vision based autonomy \cite{kayacan2018embedded,higuti2018under} through a range of Experimenta Units (EUs), or ``plots'', over which the field is divided for comparing different varieties. Field data is collected with a combination of onboard sensors including cameras and LiDAR. The angles, resolution, field-of-view, and frame rates of the cameras can be programatically customized for the desired traits. On-board software transfers the data to a remote computer, usually in the Cloud, where it is processed further. First the data is split into Experimental Units using metadata provided by the user as well as embedded GPS information. Next per-plot data is then processed through machine-vision algorithms which generate the desired output, soybean pod count in this case. The results are then associated back with the appropriate EUs in the database and presented to the user with an intuitive web-based interface. A pictorial overview of the system is provided in Figure \ref{fig:ES_pheno}

\begin{figure}
\centering
    \includegraphics[width=.8\textwidth]{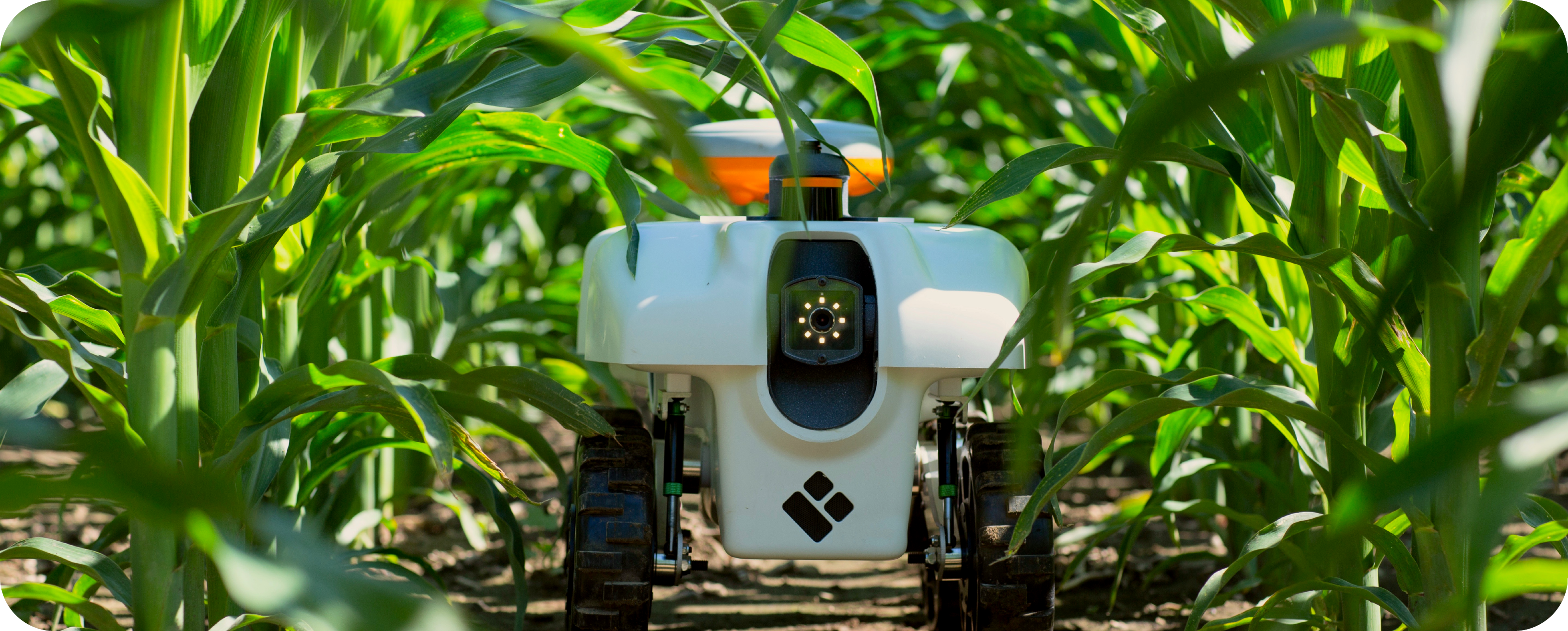}
	\caption{EarthSense TerraSentia robot is a compact robot designed to collect under-canopy traits in an autonomous manner}
	\label{fig:TS}
\end{figure}

The TerraSentia robot (\textbf{Figure} \ref{fig:TS}) is an ultra-compact (35cm wide), ultra-light (13kg), low-cost, field robot that can navigate in a variety of field conditions autonomously. TerraSentia is \emph{\textbf{(1)}} Ultra-light-weight compared to even small robotic rovers (e.g., the 17kg ClearPath Husky); \emph{\textbf{(2)}} Ultra-compact, capable of turning $360^\circ$ on the spot in 75cm rows; \emph{\textbf{(3)}} Fully autonomous under dense crop canopy where GPS does not work; and \emph{\textbf{(4)}} uniquely designed (patent pending) to not damage young plants even if it drives over them. The robot is equipped with four cameras, on the left, right, front, and top of the robot, as well as two LiDARs, RTK GPS, wheel encoders and other on-board sensors. The robot has up to 2 TB onboard storage for the data it collects. The robot can be connected to the internet using an Ethernet port, via which it can upload the data to the terrasentia.co website. On the website, the data is analyzed using custom designed machine vision algorithms which are described in Section \ref{sec:podount}. The robot is commercially manufactured by EarthSense Inc., Champaign, IL.

\subsection{Data collection and field experiments}

Data were collected October 14th - one day prior to harvesting from a field with 36 ranges and 40 columns, each column being a single row. Each plot had a unique ID associated with it from 4560 to 6000 in a serpentine path. The row spacing was 76 cm and each plot was a 1-meter long row and contained a unique experimental line. The TerraSentia robot was driven through the center between the two adjacent rows of soybean plants. The data was collected with the two side cameras fitted with ``fisheye'' lenses designed to image the entire soybean plant. Data were collected in ``Collections''-packages containing data from multiple sensors–by running the robot along an entire column at a time. Because each plot was a single row, when the robot drove in between rows, plots to the left and right of the robot were scanned simultaneously. Each collection was manually split into sub-collections corresponding to individual plots on the left and right side of the robot.

\begin{figure}
\centering
    \includegraphics[width=.8\textwidth]{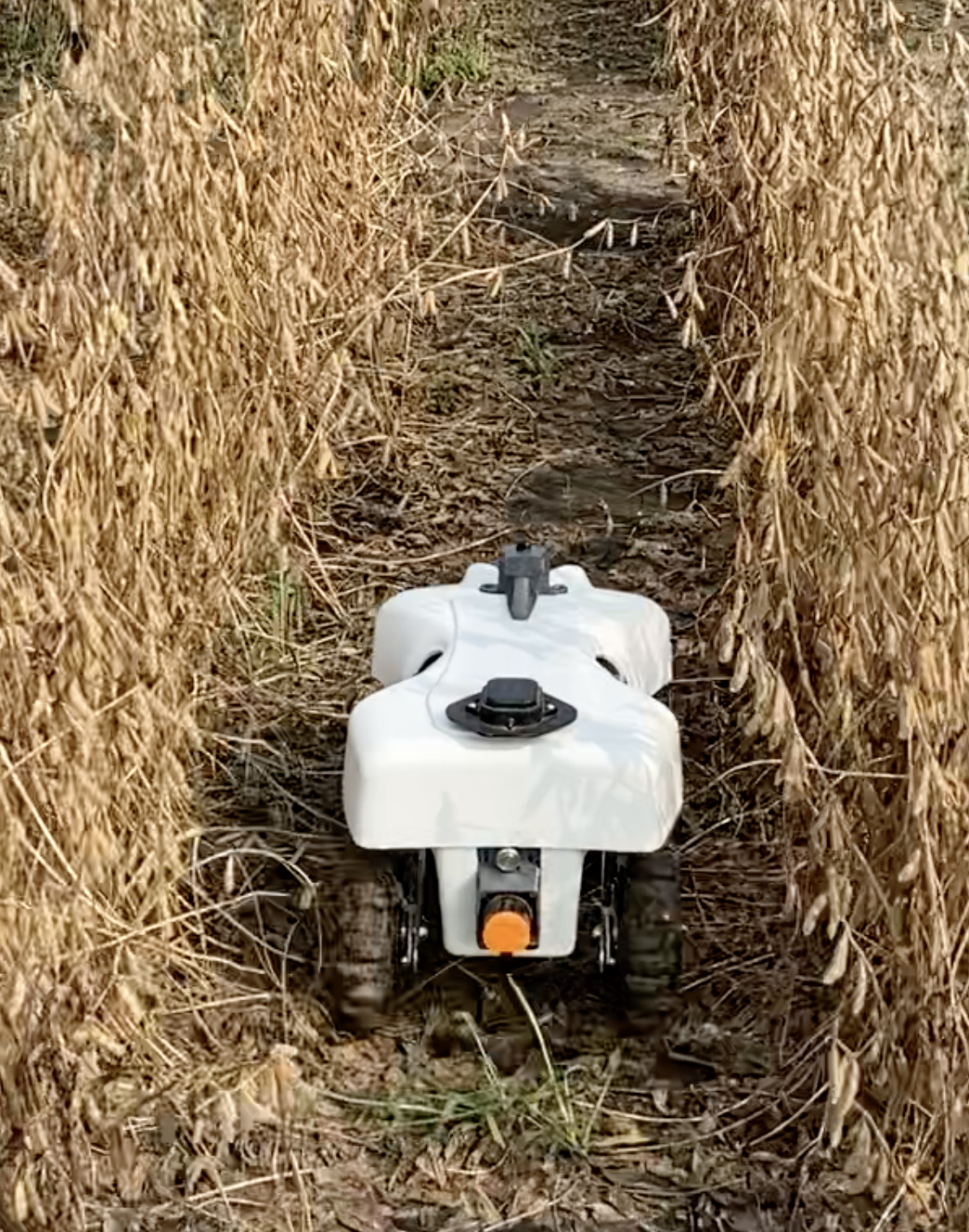}
	\caption{EarthSense TerraSentia robot gathering Soybean data}
	\label{fig:TSSoybeans}
\end{figure}

\begin{figure}[t!]
\centering
\includegraphics[scale=0.5]{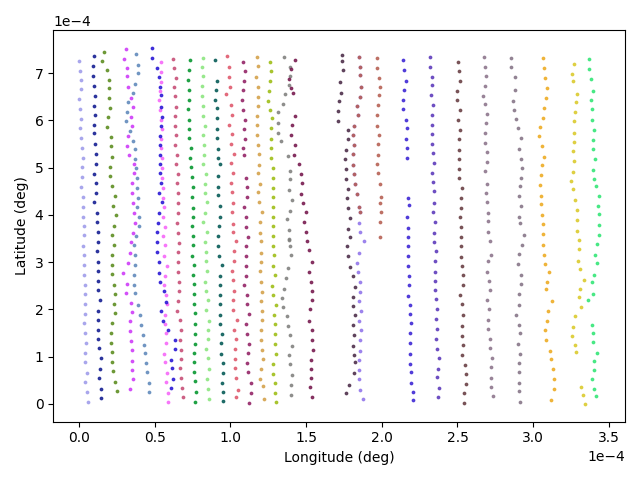}
\caption{GPS coordinates for the center of each plot in the field, extracted from the RTK GPS metadata recorded by the TerraSentia robot during data collection.}
\label{fig:GPS}
\end{figure}

The full-column collections were split into sub-collections using a plot-splitting algorithm, each corresponding to an EU. The plot-splitting algorithms can use the LiDAR data from the rear of the robot or GPS (when available) to split the data into sub-collections. The splits were manually verified to ensure correct assignment. Plot assignment was verified using the user-recorded metadata about the column number, the beginning and end plots of the collection and verified using RTK GPS metadata \ref{fig:GPS}. The splits were further verified by also checking them against the full-column correlation analysis. The pod-counting algorithm was run on each of these plots individually and the results were correlated with a of harvest yields data. The final correlational analysis described below utilizes data from all plots that have both complete data and a high degree of confidence in plot identification.

\subsection{Machine Vision based pod-count algorithm}\label{sec:podount}

\begin{figure*}[ht!]
\hspace*{\fill}
   \subfloat[\label{genworkflow}]{%
      \includegraphics[height=0.6\textwidth]{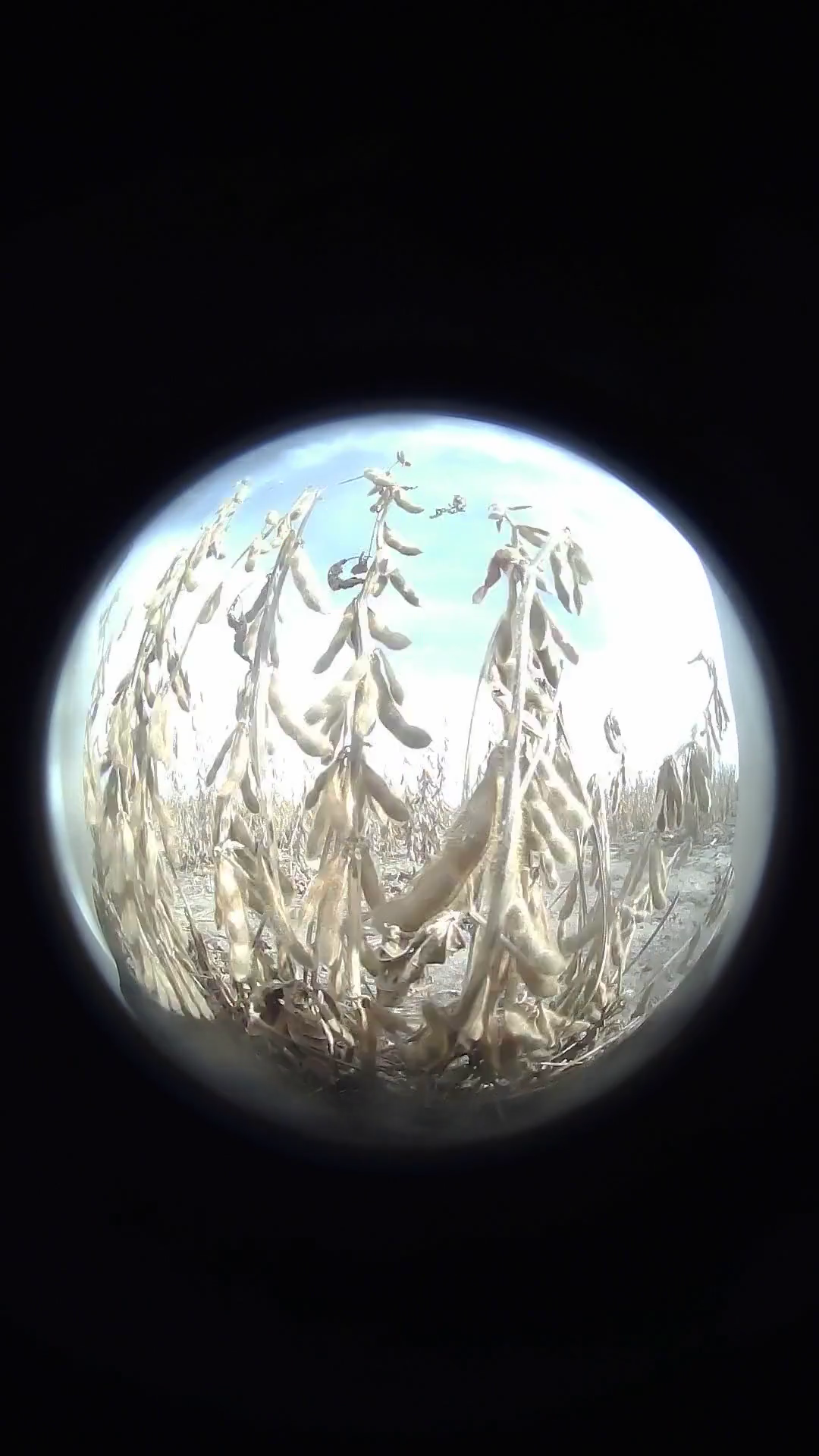}}
\hspace*{1em}
   \subfloat[\label{pyramidprocess} ]{%
      \includegraphics[height=0.6\textwidth]{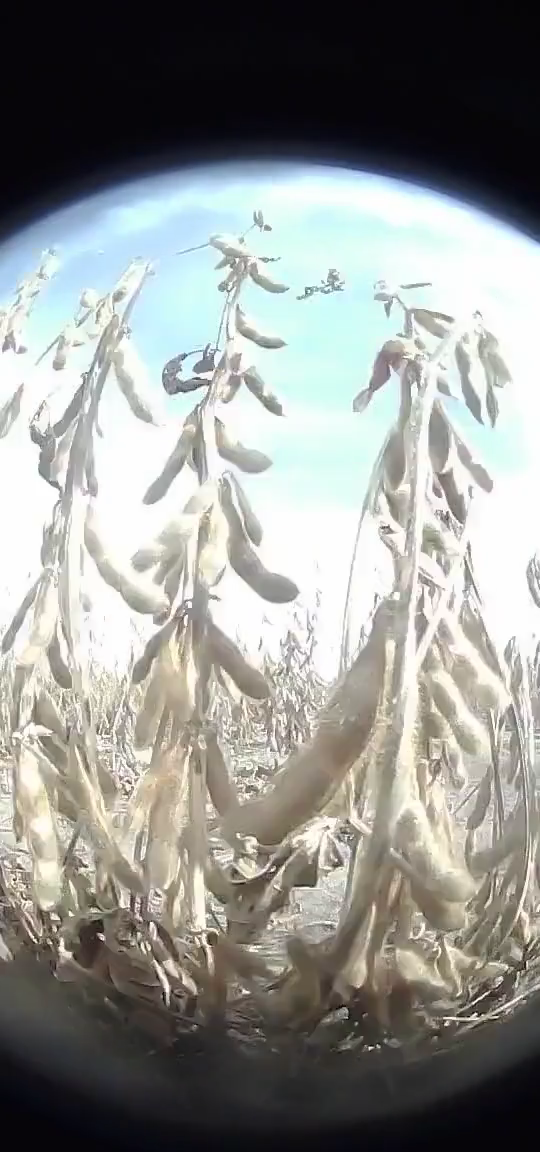}}
\hspace*{\fill}
\caption{\textbf{Left:} An example of a frame taken by our side camera with a fisheye lens. \textbf{Right:} The same frame, with preprocessing applied.}
\label{fig:fisheye}
\end{figure*}

The EarthSense pod-count algorithm was developed followed a data-detection-geometry framework. The TerraSentia robot enables a high quality data collection scheme with  video taken from cameras on the left and right of the robot, odometry, LiDAR, and GPS data, combined with user-provided metadata. In the case of soybean plants, a key challenge was optimizing the camera system to ensure that the entire soybean plants were imaged. This was achieved by using wide field-of-view lenses (also known as fisheye lenses). In pre-proessing the fisheye data was cropped on the side to focus on the plants in the center of the image. The original and cropped images are shown in figure \ref{fig:fisheye}. A proprietary detection and counting algorithm was utilized to identify and count individual soybean pods. Frames were extracted from video data such that they were equidistant in the plot, Odometry and integrated sensor fusion algorithms were utilized to minimize overlap in images and dupliacte pod counting.  The detector then identified soybean pods in each frame, and data from all frames in a plot were combined to estimate the pod-count.

\section{Results}


The algorithm's ability to predict soybean yield was assessed by characterizing the Pearson correlation between predicted pod-count and measured seed yield per plot.  This analysis reflects pod-count accuracy, since seed yield and pod counts are known to be highly correlated. Yield data and pod-counts were correlated for 1,463 plots. After the initial correlation analysis, raw data from 2-sigma outliers were examined and filtered to remove plots that had bad data, such as over-exposed images due to adverse lighting or data-loss. Outlier removal left 1,391 plots for further analysis. Finally, pod-counts for the subset of plots that had been recorded from both sides were averaged. The final correlation analysis was performed for 940 unique plots with known high-quality raw data.

\begin{figure*}[ht!]
    \subfloat[\label{soybean_analysis}]{
        \includegraphics[trim=10 0 40 0,clip,width=0.32\textwidth]{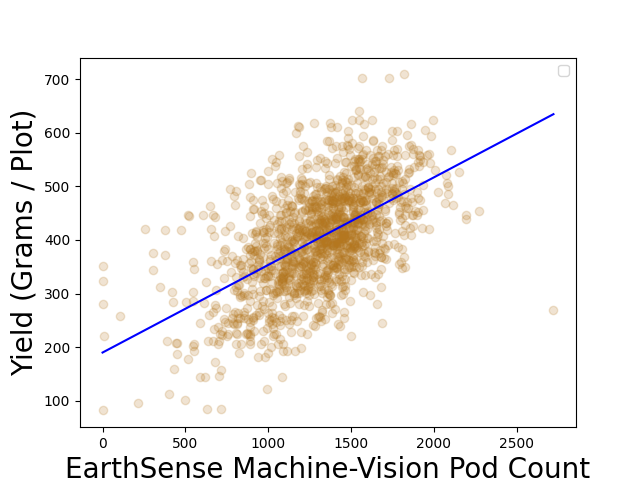}}
\hspace{0em}
    \subfloat[\label{soybean_analysis_outliers_removed}]{
        \includegraphics[trim=10 0 40 0,clip,width=0.32\textwidth]{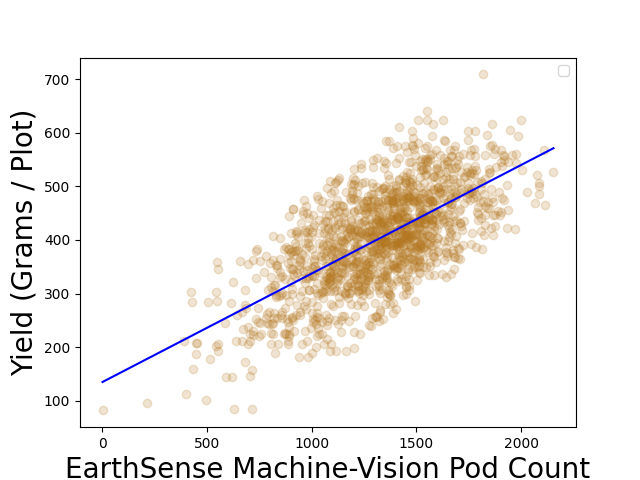}}
\hspace{0em}
    \subfloat[\label{soybean_analysis_outliers_removed_average}]{
        \includegraphics[trim=10 0 40 0,clip,width=0.32\textwidth]{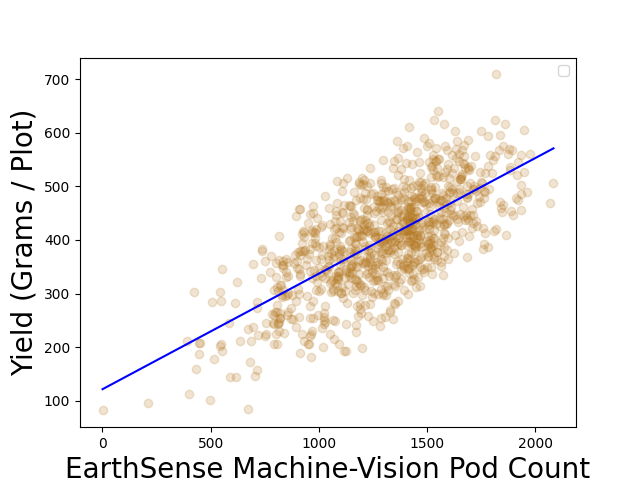}}
\caption{\textbf{Left: } Pearson correlation between TerraSentia pod-count estimates and yield measurement for the entire dataset (\textbf{N = 1463}, \textbf{r = 0.57}). \textbf{Center: } Correlation after removing outliers (\textbf{N = 1391}, \textbf{r = 0.67}). \textbf{Right: } Correlation after removing outliers and averaging algorithm counts for plots with data on from both sides (\textbf{N = 940}, \textbf{r = 0.70})}
\end{figure*}

Pearson correlation analysis shows a strong and significant correlation in all scenarios between TerraSentia pod count and manual yield measurement. For all plots, the Pearson correlation coefficient was $\textbf{r = 0.57}$ between all pod counts and their corresponding yield measurements. The subset of data after outlier filtering had a correlation of $\textbf{r = 0.67}$ between pod counts and yield measurements. Finally, the averaging of pod-counts from dual-imaged plots exhibited a Pearson correlation of $\textbf{r = 0.70}$. The correlations for the three cases are shown in Figure \ref{soybean_analysis}, Figure \ref{soybean_analysis_outliers_removed}, and Figure \ref{soybean_analysis_outliers_removed_average}.

Additionally, manual pod counts were available for 16 unique plots from a separate experiment. The correlation between TerraSentia counts and manual-counts from this experiments was $\textbf{r = 0.80}$. Filtering the single 2-sigma outlier data point improved the correlation to $\textbf{r = 0.88}$. This indicates that of TerraSentia pod-count estimates are in themselves also  accurate.


\newpage

\newpage

\section{Discussion and on-going work}

The results reported here demonstrate that high-throughput high-throughput field phenotyping for soybean pod counting is feasible. We will repeat the experiments in the 2021 season across multiple locations and varieties to validate these results over a larger dataset. Some specific directions for improvement include data quality with better sensors, improving field meta-data collection with an enhanced user-interfaces, and detector quality. With these improvements, and a larger evaluation set, we expect a greater pod-counting accuracy.

\clearpage

\bibliography{ref,daslab_all,daslab_pubs}
\bibliographystyle{unsrt}








\end{document}